\documentclass[runningheads]{llncs}
\usepackage{amsmath}
\usepackage{amssymb}
\usepackage{graphicx}
\usepackage{todonotes}
\usepackage{booktabs}
\usepackage[english]{babel}
\usepackage{float}
\usepackage[utf8x]{inputenc}
\usepackage{calc}
\usepackage{verbatim}

\usepackage{graphicx,scrpage2,placeins,lscape,
  colortbl,longtable,caption,footmisc,
  multirow}
\usepackage{hyperref}

\definecolor{lime}{HTML}{A6CE39}
\DeclareRobustCommand{\orcidicon}{%
	\begin{tikzpicture}
	\draw[lime, fill=lime] (0,0) 
	circle [radius=0.16] 
	node[white] {{\fontfamily{qag}\selectfont \tiny ID}};
	\draw[white, fill=white] (-0.0625,0.095) 
	circle [radius=0.007];
	\end{tikzpicture}
	\hspace{-2mm}
}

\foreach \x in {A, ..., Z}{%
	\expandafter\xdef\csname orcid\x\endcsname{\noexpand\href{https://orcid.org/\csname orcidauthor\x\endcsname}{\noexpand\orcidicon}}
}

\usepackage{color}
\usepackage{tabularx}
\usepackage{hhline}
\usepackage{array}

\newcolumntype{M}[1]{>{\centering\arraybackslash}m{#1}}

\newcommand{\ie}[0]{i.\,e.\ }
\newcommand{\ppm}{PBPM} %
\newcommand{\magic}[0]{PrBPM} %

\begin{document}
\title{Prescriptive Business Process Monitoring for Recommending Next Best Actions}
\titlerunning{Recommending next best actions}
\author{Sven Weinzierl\orcidA{} \and
Sebastian Dunzer\orcidB{} \and
Sandra Zilker\orcidC{} \and \\
Martin Matzner\orcidD{}}
\authorrunning{S. Weinzierl et al.}
\institute{
Friedrich-Alexander-Universit{\"a}t Erlangen-N{\"u}rnberg, F{\"u}rther Stra{\ss}e 248, N{\"u}rnberg, Germany
\email{\{sven.weinzierl, 
sebastian.dunzer, 
sandra.zilker, 
martin.matzner\}@fau.de}\\
}

\maketitle              %
\begin{abstract}
Predictive business process monitoring (\ppm{}) techniques predict future process behaviour based on historical event log data to improve operational business processes. 
Concerning the next activity prediction, recent \ppm{} techniques use state-of-the-art deep neural networks (DNNs) to learn predictive models for producing more accurate predictions in running process instances.
Even though organisations measure process performance by key performance indicators (KPIs), the DNN's learning procedure is not directly affected by them. Therefore, the resulting next most likely activity predictions can be less beneficial in practice.
Prescriptive business process monitoring (\magic) approaches assess predictions regarding their impact on the process performance (typically measured by KPIs) to prevent undesired process activities by raising alarms or recommending actions.
However, none of these approaches recommends actual process activities as actions that are optimised according to a given KPI.  
We present a \magic{} technique that transforms the next most likely activities into the next best actions regarding a given KPI. 
Thereby, our technique uses business process simulation to ensure the control-flow conformance of the recommended actions.
Based on our evaluation with two real-life event logs, we show that our technique's next best actions can outperform next activity predictions regarding the optimisation of a KPI and the distance from the actual process instances.  
\end{abstract}

\section{Introduction}
Predictive business process monitoring (\ppm) techniques predict future process behaviour to improve operational business processes~\cite{maggi.2014}. %
A \ppm{} technique constructs predictive models from historical event log data~\cite{marquez.2017} to tackle different prediction tasks like predicting next activities, %
process outcomes %
or remaining time~\cite{di.2018}. %
Concerning the next activity prediction, recent \ppm{} techniques use state-of-the-art deep neural networks (DNNs) to learn predictive models for producing more accurate predictions in running process instances~\cite{weinzierl.2020}. 
DNNs belong to the class of deep-learning (DL) algorithms. DL is a subarea of machine learning (ML) that identifies intricate structures in high-dimensional data through multi-representation learning~\cite{lecun.2015}.
After learning, models can predict the next most likely activity of running process instances.

However, providing the next most likely activity does not necessarily support process stakeholders in process executions~\cite{teinemaa.2018}. 
Organisations measure the performance of processes through key performance indicators (KPIs) in regard to three dimensions: time, cost and quality~\cite{vanderAalst.2016}.
Recent PBPM techniques rely on state-of-the-art DNNs that can only learn predictive models from event log data. 
Even though an event log can include KPI information, it does not directly affect such an algorithm's learning procedure unless a KPI is the (single) learning target itself.
As a consequence, the learned models can output next activity predictions, which are less beneficial for process stakeholders.    

Some works tackled this problem with prescriptive business process monitoring (\magic) approaches.
\magic{} approaches assess predictions regarding their impact on the process performance -- typically measured by KPIs -- to prevent undesired activities~\cite{teinemaa.2018}.
To achieve that, existing approaches generate alarms 
~\cite{teinemaa.2018,fahrenkrog.2019,metzger2017predictive,metzger2019dl} or recommend actions~\cite{conforti.2013,groger.2014}.
However, none of these approaches recommends next best actions in the form of process activities that are optimised regarding a given KPI for running processes. 
In our case, \emph{best} refers to a KPI's optimal value regarding the future course of a process instance.
Additionally, the next best actions, which depend on next activity predictions and prediction of a particular KPI, might obscure the actual business process. Therefore, transforming methods should check whether a recommended action is conform regarding a process description.
\vspace{-0.5cm}
\begin{figure}[htb]
\centering
\includegraphics[width=7.5cm]{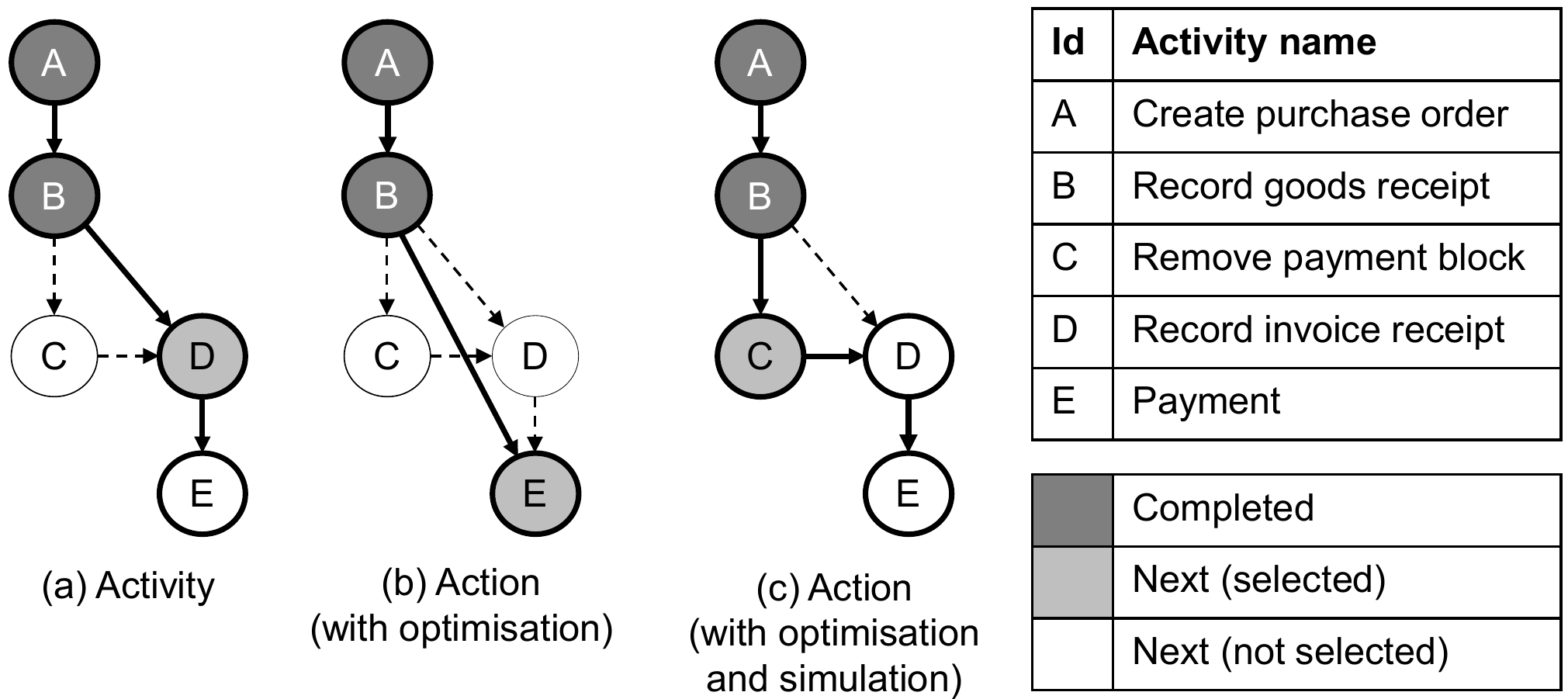}
\caption{A next activity prediction vs. a next best action recommendation.} 
\label{fig:predact}
\end{figure}
\vspace{-0.7cm}

Given a running process instance of a purchase order handling process after finishing the first two activities, a DNN model predicts the next most likely activity D (cf. (a) in Fig.~\ref{fig:predact}). Additionally, the KPI \textit{time} is of interest and the the first two activities A and B take each $1$ hour, the predicted activity D takes $2$ hours and the last activity E takes $2$ hours. In sum, the complete process instance takes $6$ hours and a deadline of $5$ hours -- that exists due to a general agreement -- is exceeded.
In contrast, the recommended action with optimisation can be the activity E (cf. (b) in Fig.~\ref{fig:predact}). Even though the complete process instance takes $4$ hours, the mandatory activity D is skipped.  
With an additional simulation, the activity ``Remove payment block" can be recommended (cf. (c) in Fig.~\ref{fig:predact}) taking $1$ hour. Afterwards, the activities D and E are followed with a duration of $2$ hours and $1$ hour. Here, the activity E takes $1$ hour instead of $2$ hours since the payment block is already removed. Thus, the complete process instance takes $5$ hours, and the deadline is met. 

In this paper, we provide a \magic{} technique for recommending the next best actions depending on a KPI. Thereby, it conducts a business process simulation (BPS) to remain within the allowed control-flow. 
To reach our research objective, we develop a \magic{} technique and evaluate its actions regarding the optimisation of a KPI and the distance from ground truth process instances with two real-life event logs. 

This paper is an extended and revised version of a research-and-progress paper~\cite{weinzierl.2020c}. Additionally, it includes a BPS and an evaluation with two real-life logs.  
The paper is structured as follows:
Sec.~\ref{sec:background} presents the required background for our~\magic{}~technique.
Sec.~\ref{sec:artifact} introduces the design of our \magic{} technique for recommending next best actions.
Further, we evaluate our technique in Sec.~\ref{sec:eval}. 
Sec.~\ref{sec:discussion}~provides a discussion.
The paper concludes with a summary and an outlook on future work in Sec.~\ref{sec:conclusion}.

\section{Background}
\label{sec:background}
\subsection{Preliminaries}
\ppm\ or PrBPM techniques require event log data. We adapt definitions by Polato et al.~\cite{polato.2014} to formally describe the terms \textit{event}, \textit{trace}, \textit{event log}, \textit{prefix} and \textit{suffix}.   
In the following, $\mathcal{A}$ is the set of process activities, $\mathcal{C}$ is the set of process instances~(cases), and $C$ is the set of case ids with the bijective projection $id : C \to \mathcal{C}$, and $\mathcal{T}$ is the set of timestamps.
To address time, a process instance $c \in \mathcal{C}$ contains all past and future events, while events in a trace $\sigma_c$ of $c$ contain all events up to the currently available time instant.
$\mathcal{E}= \mathcal{A} \times C \times \mathcal{T}$ is the event universe.

\begin{definition}[Event] An event $e \in \mathcal{E}$
  is a tuple $e=(a,c,t)$,
  where $a \in \mathcal{A}$
  is the process activity,
  $c \in C$ is the case id,
  and $t \in \mathcal{T}$ is its timestamp.
  Given an event $e$,
  we define the projection functions
  $F_{p}=\{f_{a}, f_{c}, f_{t}\}$: $f_{a}: e  \to a, f_{c}: e \to c, \text{and}\ f_{t}: e \to t$.
\end{definition}
\begin{definition}[Trace] A trace is a %
  sequence $\sigma_{c} = \langle e_{1}, \dots, e_{\vert \sigma_{c} \vert} \rangle \in \mathcal{E}^*$ of events, such that $f_{c}(e_{i}) = f_{c}(e_{j}) \wedge f_{t}(e_i) \leq f_{t}(e_j)$ for $1 \leq i < j \leq \vert \sigma_{c} \vert$. Note a trace $\sigma_{c}$ of process instance $c$ can be considered as a process instance $\sigma_{c}$. 
  \end{definition}
  
\begin{definition}[Event log]
  An event log $\mathcal{L}_\tau$
  for a time instant $\tau$
  is a set of traces,
  such that 
  $\forall \sigma{}_c \in \mathcal{L}_\tau\,.\,\exists c \in \mathcal{C}\,.\, (\forall e \in \sigma_c\,.\,id(f_c(e)) = c) \wedge (\forall e \in \sigma_{c}\,.\,f_t(e) \leq \tau)$,
  \ie all events of the observed cases that already happened.
  
\end{definition}

\begin{definition}[Prefix, suffix of a trace]
  Given a trace $\sigma_{c}= \langle e_{1},..,e_{k},.., e_{n} \rangle$,
  the prefix of length $k$ 
  is $hd^k(\sigma_{c})= \langle e_{1},..,e_{k} \rangle$, and
  the suffix of length $k$ is $tl^k(\sigma_{c})=\langle e_{k+1},.., e_{n} \rangle$, with $1 \leq k < n$. 
\end{definition}

\subsection{Long short-term memory neural networks}
Our \magic{} technique transforms next activity predictions into the next best actions. Thus, next activity predictions are the basis for our \magic{} technique.  
To predict next activities, we use a ``vanilla", i.e. basic, long short-term memory network (LSTM)~\cite{hochreiter.1997} because most of the PBPM techniques for predicting next activities rely on this DNN architecture~\cite{weinzierl.2020b}. LSTMs belong to the class of recurrent neural networks (RNNs)~\cite{lecun.2015} and are designed to handle temporal dependencies in sequential prediction problems~\cite{bengio.1994}.
In general, consists of three layers: an input layer (receiving data input), a hidden layer (i.e. an LSTM layer with an LSTM cell) and an output layer (providing predictions).  

An LSTM cell uses four gates to manage its memory over time to avoid the problem of gradient exploding/vanishing in the case of longer sequences~\cite{bengio.1994}.
First, a forget gate that determines how much of the previous memory is kept. Second, an input gate controls how much new information is stored into memory. Third, a gate gate or candidate memory that defines how much information is stored into memory. Fourth, an output gate that determines how much information is read out of the memory.

To learn an LSTM's parameters, a loss function (e.g. the cross-entropy loss for classification) is defined on a data point (i.e. prediction and label) and measures the penalty. Additionally, a cost function in its basic form calculates the sum of loss functions over the training set. 
The LSTM's parameters are updated iteratively via a gradient descent algorithm (e.g. stochastic gradient descent), in that, the gradient of the cost function is computed by backpropagation through time~\cite{rumelhart.1986}. After learning the parameters, an LSTM model with adjusted parameter values exists.

\subsection{Business process simulation}
Actions optimised according to a KPI can be not conform to the process control-flow.
Thus, suggesting process-conform actions to process stakeholders is essential.
Consequently, we add control-flow knowledge to our \magic{} technique with formal process models.

A well-known approach to assess the quality of process executions is business process simulation (BPS).
Several approaches examine processes and their variants regarding compliance or performance with BPS~\cite{centobelli.2015,redlich.2012}.
We refer to discrete-event-driven BPS~\cite{tumay.1996}. Here, simulation models formally contain discrete events which are interrelated via process semantics.

BPS usually delivers its insights to users~\cite{rosenthal.2018}.
Unlike existing approaches, such as~\cite{wynn.2008,rozinat.2009}, we use the simulation results to process the predictions of an LSTM.
Thus, we use discrete-event-driven~\cite{tumay.1996} short-term simulation~\cite{rozinat.2009} as a boundary measure to ensure that the DNN-based next best action makes sense from a control-flow perspective. Alike Rozinat et al.~\cite{rozinat.2009}, our simulation starts from a non-empty process state to aid in recommending the next best action from the current state on.

\section{A PrBPM technique for recommending next best actions}
\label{sec:artifact}
Our PrBPM technique transforms next activity predictions into the next best actions. %
The technique consists of an \textit{offline} and an \textit{online component}.
In the offline component, it learns a DNN for predicting next activities and values of a KPI. %
In the online component, the next best actions are recommended based on the next activity and KPI value predictions. 
\subsection{Offline component}
\begin{figure}[htb]
\centering
\includegraphics[width=11cm]{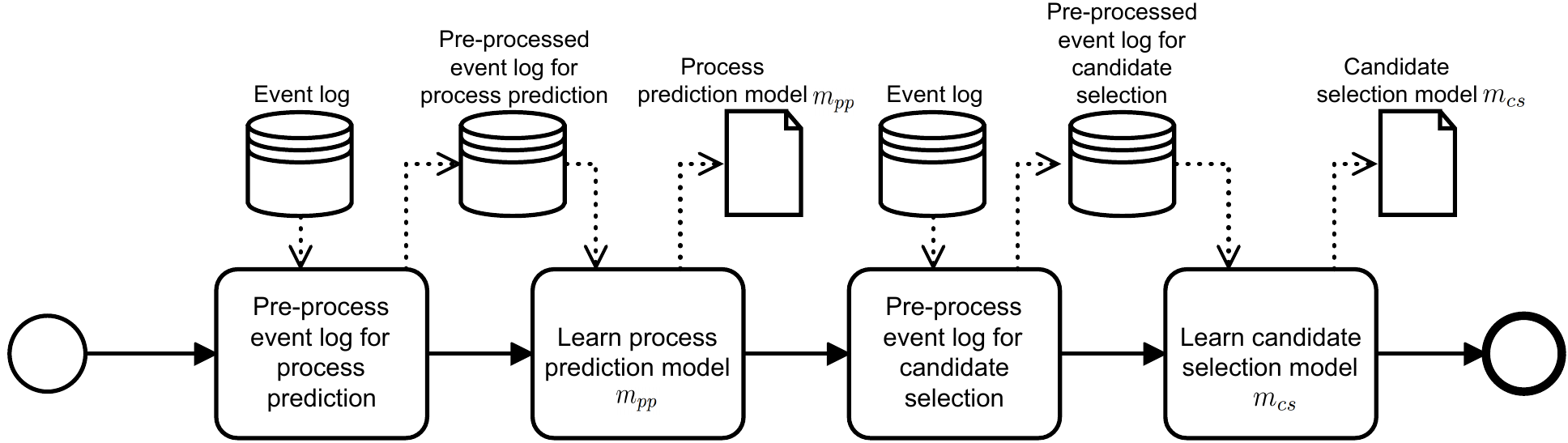}
\caption{Four-step offline component scheme with the two models $m_{pp}$ and $m_{cs}$.} 
\label{fig:rec_offline}
\end{figure}

The offline component receives as input an event log $\mathcal{L}_\tau$, and outputs the two ML models $m_{pp}$ and $m_{cs}$.
While $m_{pp}$ (process prediction model) predicts next activities and a KPI value related to next activities, 
$m_{cs}$ (candidate selection model) selects a fix set of suffix candidates. 
The technique learns both models from individually pre-processed versions of $\mathcal{L}_\tau$.
Fig.~\ref{fig:rec_offline}~visualises the steps of the offline component.
\vspace{-0.5cm}
\vspace{-0.5cm}
In the following, we describe the steps of the offline component based on the exemplary finished process instance $\sigma_{1}^f$, as represented in (\ref{eq:example}). The last attribute per event is the KPI; here the defined costs for executing an activity. 

\begin{equation}
    \begin{split}
      \sigma_{1}^f=\langle&\text{(1, ``Create Application", 2011-09-30 16:20:00, 0),}\\
        &\text{(1, ``Concept", 2011-09-30 17:30:00, 10),}\\
        &\text{(1, ``Accepted", 2011-09-30 18:50:00, 20),}\\
        &\text{(1, ``Validating", 2011-09-30 19:10:00, 40)}
      \rangle.
    \end{split}
  \label{eq:example}
\end{equation}

\textbf{Pre-process event log for process prediction.}
$m_{pp}$ is a multi-task DNN for predicting next activities and a KPI value at each time step of a running process instance. For $m_{pp}$, the pre-processing of $\mathcal{L}_\tau$ comprises four steps. 
First, to determine the end of each process instance in $\mathcal{L}_\tau$, it adds a termination event to the end of each process instance. So, for $\sigma_{1}^{f}$, as represented in (\ref{eq:example}), we add the event $\text{(1, ``End", 2011-09-30 19:10:00, 0)}$ after the fourth event with the activity name ``Validating". Additionally, for termination events, we always overtake the timestamp value of the previous event and set the value of the KPI to $0$.  
Second, it onehot-encodes all activity 
names in the process instances as numeric values (cf. (\ref{eq:example_2}) for $\sigma_{1}^f$ including the termination event's activity).

\begin{equation}
    \begin{split}
      \sigma_{1}^f=\langle
        (0, 0, 0, 0, 1),
        (0, 0, 0, 1, 0),
        (\dots),
        (0, 1, 0, 0, 0),
        (1, 0, 0, 0, 0)
      \rangle.
    \end{split}
  \label{eq:example_2}
\end{equation}

This step is necessary since LSTMs, as used in this paper, use a gradient descent optimisation algorithm to learn the network's parameters. 
Third, it crops prefixes out of process instances by using the function $hd^k()$. For instance, a prefix with size three of $\sigma_{1}^f$ is:

\begin{equation}
    \begin{split}
      hk^3(\sigma_{1}^f)=\langle
        \text{(0, 0, 0, 0, 1),}&\text{(0, 0, 0, 1, 0),}
        \text{(0, 0, 1, 0, 0)}
      \rangle.
    \end{split}
  \label{eq:example_3}
\end{equation}

Lastly, it transforms the cropped input data into a three-order tensor (prefixes, time steps and attributes). Additionally, $m_{pp}$ needs two label structures for parameter learning. First, for the onehot-encoded next activities, a two-dimensional label matrix is required. Second, if the KPI values related to the next activities are scaled numerically, a one-dimensional label vector is required. If the values are scaled categorically, a two-dimensional label matrix is needed.

\textbf{Create process prediction model.}
$m_{pp}$ is a multi-task DNN.
The model's architecture follows the work of Tax et al.~\cite{tax.2017}.  
The input layer of $m_{pp}$ receives the data and transfers it to the first hidden layer. The first hidden layer is followed by two branches. Each branch refers to a prediction task and consists of two layers, a hidden layer and an output layer. The output layer of the upper branch realises next activity predictions, whereas the lower creates KPI value predictions. 
Depending on the KPI value's scaling (i.e. numerical or categorical), the lower branch solves either a regression or classification problem.
Each hidden layer of $m_{pp}$ is an LSTM layer with an LSTM cell.

\textbf{Pre-process event log for candidate selection.}
$m_{cs}$ is a nearest-neighbour-based ML algorithm for finding suffixes ``similar" to predicted suffixes. 
For $m_{cs}$, the pre-processing of $\mathcal{L}_\tau$ consists of three steps.
First, it ordinal-encodes all activity names in numerical values. For example, the ordinal-encoded representation of $\sigma_{1}^f$, as depicted in (\ref{eq:example}) including the termination event's activity, is $\langle (1), (2), (3), (4), (5) \rangle$.
Second, it crops suffixes out of process instances through the function $tl^k(\cdot)$. For instance, the suffix with size three of $\sigma_{1}^f$ ($lt^3(\sigma_{1}^f))$ is $\langle (4), (5) \rangle$. 
Lastly, the cropped input data is transformed into a two-dimensional matrix (suffixes and attributes). 

\textbf{Create candidate selection model.}
$m_{cs}$ is a nearest-neighbour-based ML algorithm. 
It retrieves $k$ suffixes ``nearest" to a suffix predicted for a given prefix (i.e. a running process instance at a certain time step). 
The technique learns the model $m_{cs}$ based on all suffixes cropped out of $\mathcal{L}_\tau$.

\subsection{Online component}
\begin{figure}[htb]
\centering
\includegraphics[width=\textwidth]{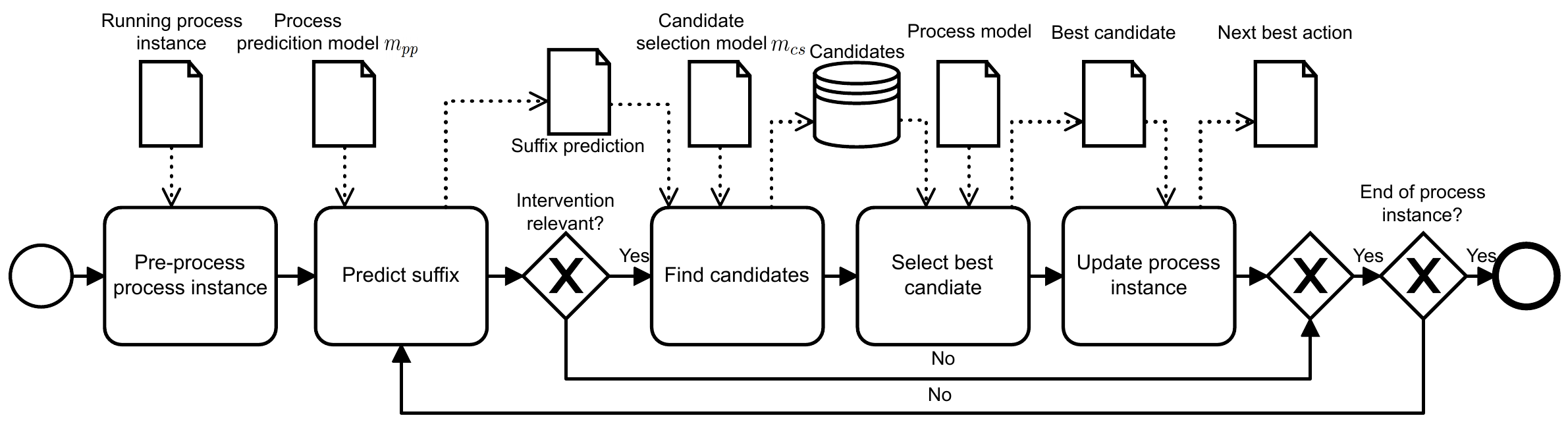}
\caption{Five-step online component scheme with the two models $m_{pp}$ and $m_{cs}$.} 
\label{fig:rec_phase}
\end{figure}

The online component receives as input a new process instance, and the two trained predictive models $m_{pp}$ and $m_{cs}$. It consists of five steps~(see~Fig.~\ref{fig:rec_phase}), and outputs next best actions.
After pre-processing (first step) of the running process instance, a suffix of next activities and its KPI values are predicted (second step) by applying $m_{pp}$. 
The second step is followed by the condition, whether the sum of the KPI values of the suffix and the respective prefix exceeds a threshold or not. If the threshold is exceeded, the predicted suffix of activities is transferred from the second to the third step (i.e. find candidates) and the procedure for generating next best actions starts. Otherwise, it provides the next most likely activity.
To find a set of suffix candidates, the technique loads $m_{cs}$ from the offline component. Subsequently, it selects the best candidate from this set depending on the KPI and concerning BPS. 
Finally, the first activity of the selected suffix represents the best action and is concatenated to the prefix of activities (i.e. running process instance at a certain time step). 
If the best action is the end of the process instance, the procedure ends. Otherwise, the procedure continues and predicts the suffix of the new prefix.    
\vspace{-0.5cm}
\vspace{-0.5cm}
In the following, we detail the online component's five steps. Thereby, we refer to the running process instance~$\sigma_{2}^r$, for that the second event has just finished.   

\begin{equation}
    \begin{split}
      \sigma_{2}^r=\langle&\text{(2, ``Create Application", 2012-09-30 18:00:00, 20),}\\
        &\text{(2, ``Concept", 2012-09-30 18:30:00, 20)}\rangle.
    \end{split}
  \label{eq:example_4}
\end{equation}

\textbf{Pre-process process instance.} To predict the suffix of the running process instance with $m_{pp}$, we onehot-encode all activity names in numerical values and transform the output into a third-order tensor.   

\textbf{Predict suffix.}
Based on a running process instance (prefix), $m_{pp}$ predicts the next sequence of activities and the KPI values.
To get the complete suffix, we apply $m_{pp}$ repeatedly.
Afterwards, the technique calculates the sum of the KPI values over the activities of the complete process instance consisting of the prefix and its predicted suffix. 
For instance, if the prefix of a process instance is $\sigma_{2}^r$, one potential suffix is: 

\begin{equation}
    \begin{split}
      s_{\sigma_{2}^r}=\langle
        \text{(``Accepted", 20),}
        \text{(``Validating", 40),}
        \text{(``End", 10)}
      \rangle.
    \end{split}
  \label{eq:example_5}
\end{equation}

For a better intuition, we omit the suffixes' encoding in the online component. The values $20$, $40$ and $10$ assigned to the events are KPI values (e.g. cost values) predicted by $m_{pp}$.    
In line with Tax et al.~\cite{tax.2017}, we do not perform the suffix prediction for prefixes with size $\leq 1$ since the amount of activity values is insufficient. 
After predicting the suffix, the total costs of $\sigma_{2}^r$ are $110$. 
To start the procedure for recommending the next best actions, the total KPI value of an instance has to exceed a threshold value $t$. The value of $t$ can be defined by domain experts or derived from the event log (e.g. average costs of process instances).
Regarding $\sigma_{2}^r$, the procedure starts because we assume $t=100$ ($110>t$). 

\textbf{Find candidates.}
Second, for the predicted suffix, $m_{cs}$ from the offline component reveals a set of alternatives with a meaningful control-flow. 
For example, $m_{cs}$ ($k=3$) selects based on $s_{\sigma_{2}^r}$ the following three suffix alternatives:

\begin{equation}
    \begin{split}
      m_{cs}(s_{\sigma_{2}^r})=[
      &\langle
        \text{(``End", 10)}
      \rangle,\\
      &\langle
        \text{(``Accepted", 20),}
        \text{(``Validating", 40),}
        \text{(``End", 10)}
      \rangle,\\
      &\langle
        \text{(``Validating", 20),}
        \text{(``Accepted", 10),}
        \text{(``Validating", 10),}\\
        &\text{(``End", 10)}
      \rangle].
    \end{split}
  \label{eq:example_6}
\end{equation}

In (\ref{eq:example_6}), the first and the third suffix result in total costs ($50$ and $90$) falling below $t$.

\textbf{Select the best candidate.}
In the third step, we select the next best action from the set of possible suffix candidates.
We sort the suffixes by the KPI value. Thus, the first suffix is the best one in regard to the KPI.
To incorporate control-flow knowledge, a simulation model checks the resulting instance.
Thereby, we reduce the risk of prescribing nonsensical actions.
The simulation uses a formal process model to retrieve specific process semantics.
The simulation produces the current process state from the prefix and the process model. 
If the prefix does not comply with the process model, the simulation aborts the suffix selection for the prefix and immediately recommends an intervention.
Otherwise, we check the $k$ selected suffixes whether they comply with the process model in the simulation from the current process state on. 
If a candidate suffix fails the simulation, 
our technique omits it in the selection.
However, when all suffix candidates infringe the simulation, the technique assumes the predicted next activity as the best action candidate.
Concerning the candidate set from (\ref{eq:example_6}), the best candidate is suffix three since it does not infringe the simulation model. 

\textbf{Update process instance.}
To evaluate our technique, we assume that a process stakeholder performs in each case the recommended action. Thus, if the best suffix candidate exists, the activity (representing the next best action) and the KPI value of the first event are concatenated to the running process instance (i.e. prefix).
After the update,~$\sigma_{2}^r$ comprises three events, as depicted in~(\ref{eq:example_7}).   

\begin{equation}
    \begin{split}
      \sigma_{2}^r=\langle
      &\text{(2, ``Create Application", 2012-09-30 18:00:00, 20),}\\
      &\text{(2, ``Concept", 2012-09-30 18:30:00, 20),}\\
      &\text{(2, ``\textbf{Validating}", --, \textbf{20})}\rangle.
    \end{split}
  \label{eq:example_7}
\end{equation}

The technique repeats the complete procedure until the termination event is reached.

\section{Evaluation}
\label{sec:eval}
We provide an evaluation regarding our PrPBM technique's optimisation of a KPI and the distance from ground truth process instances. For that, we developed a prototype that recommends next best actions depending on the KPI \textit{throughput time} and concerning a process simulation realised with DCR graphs. We compare our results to a representative baseline~\cite{tax.2017} for two event logs.  

\subsection{Event logs}
First, we use the~\textit{helpdesk}\footnote{https://data.mendeley.com/datasets/39bp3vv62t/1.}~event log containing data from an Italian software company's ticketing management process. 
It includes $21,348$ events, $4,580$ process instances, $226$ process instance variants and $14$ activities.
Second, we include the \textit{bpi2019}\footnote{https://data.4tu.nl/repository/uuid:a7ce5c55-03a7-4583-b855-98b86e1a2b07.} event log from the BPI challenge 2019, provided by a company for paints and coatings.
It depicts a purchase order handling processes. For this event log, we only considered a random 10\%-sampling with sequences of $30$ events or shorter, due to the high computation effort. 
It includes $101,714$ events, $24,900$ process instances, $3,255$ process instance variants and $32$ activities. 

\subsection{Process models}
\begin{figure}[htb!]
\centering
\includegraphics[width=.55\textwidth]{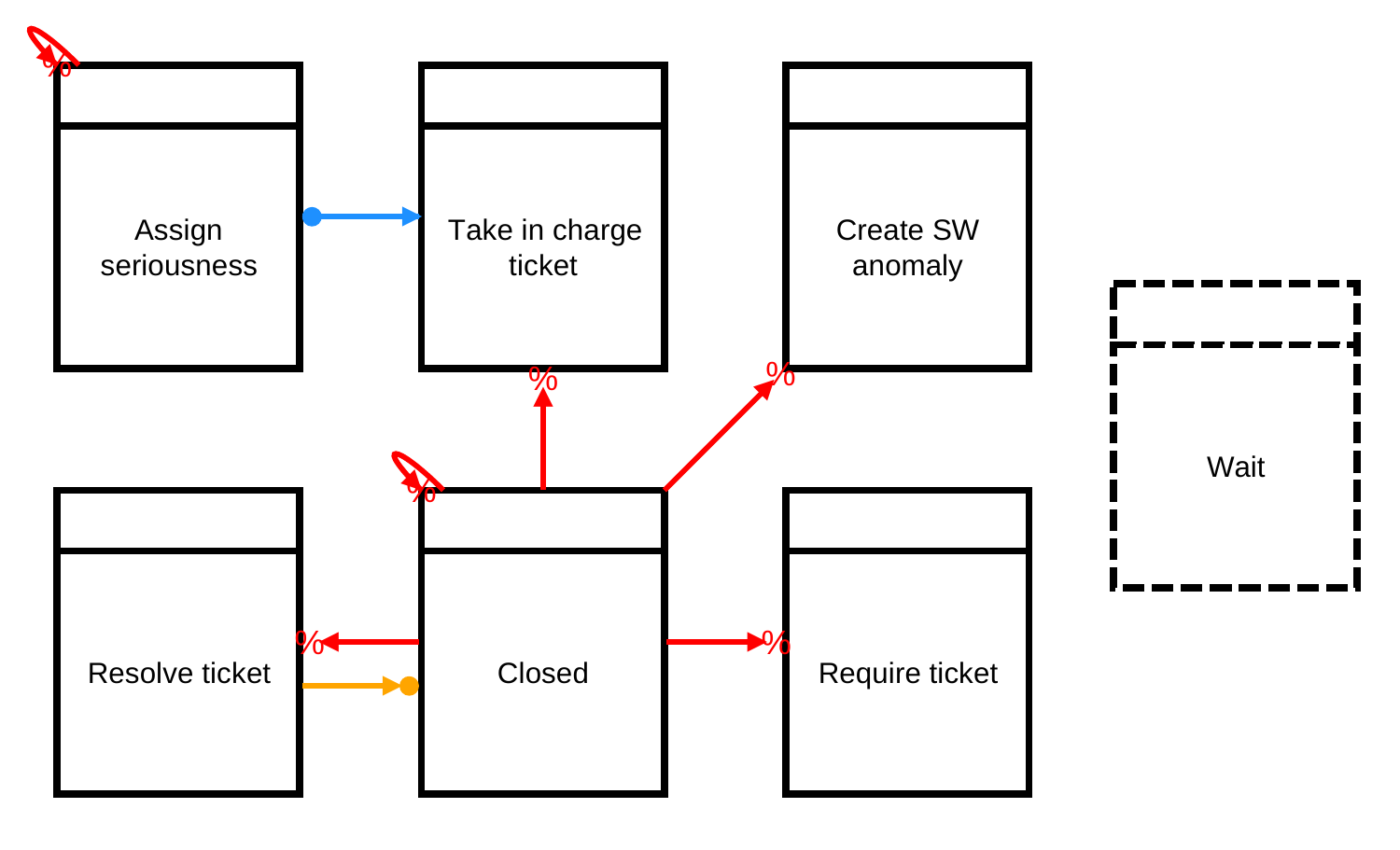}
\caption{DCR graph for the \textit{helpdesk} event log.} 
\label{fig:dcr_hd}
\end{figure}

We used DCR graphs as models for the BPS in our technique's best candidate selection. In Fig.~\ref{fig:dcr_hd}, we present the DCR graph for the \textit{helpdesk} event log. The three most important constraints are the following. First, after ``Closed" the other activities should not happen. Second, if ``Assign seriousness" occurs, someone must take over the responsibility. Third, before a ticket is closed, ``Resolve ticket" must occur.

\begin{figure}[htb!]
\centering
\includegraphics[width=\textwidth]{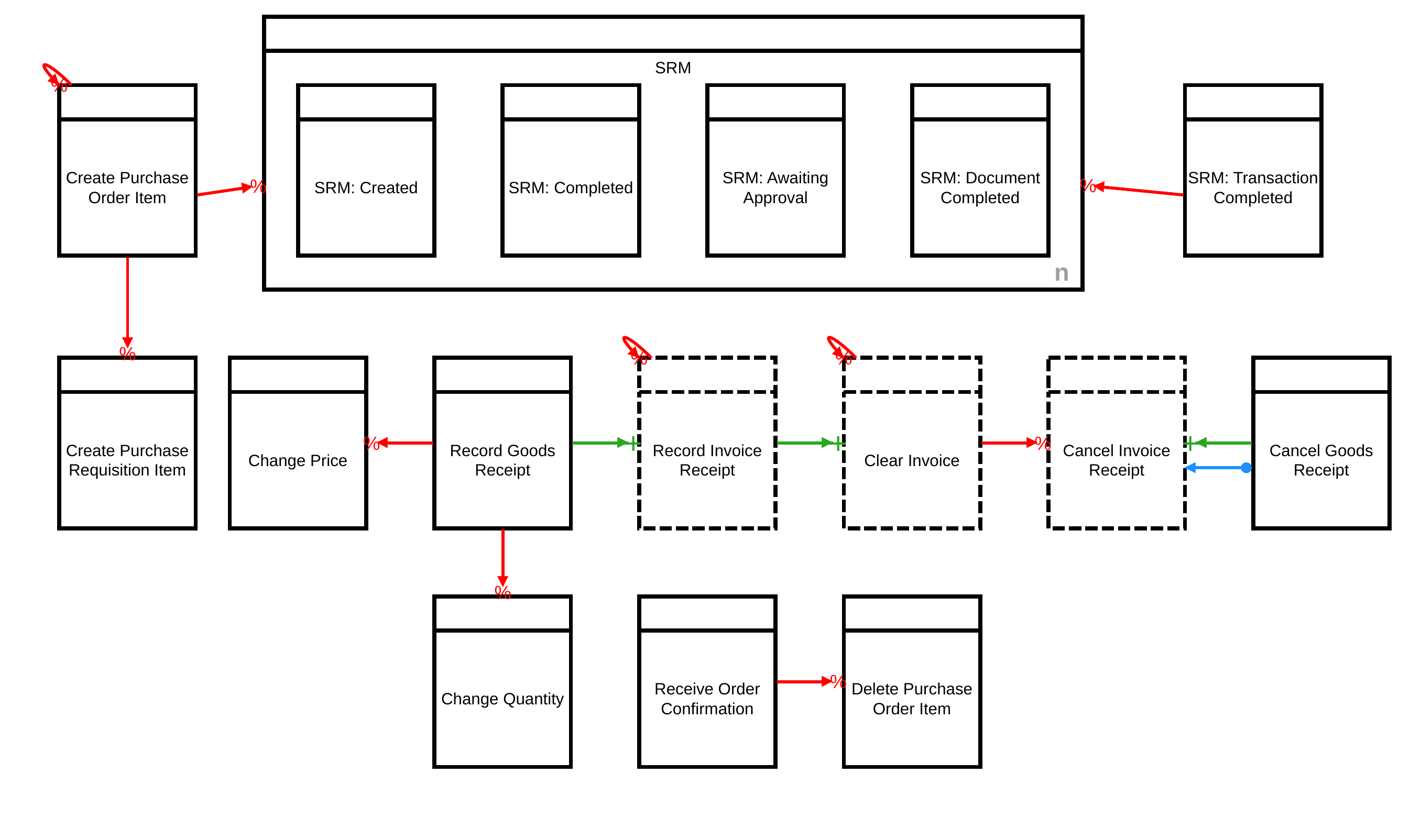}
\caption{DCR graph for the \textit{bpi2019} event log.} 
\label{fig:dcr_bpi}
\end{figure}

Fig.~\ref{fig:dcr_bpi} shows the DCR graph for the \textit{bpi2019} event log.  
The three most essential constraints are the following. First, ``Create Purchase Order Item" may only happen once per order. Second, After the goods were received, ``Change Quantity" and ``Change price" should not occur. Third, ``Record Goods Receipt", ``Record Invoice Receipt" and ``Clear Invoice" must eventually follow each other. 

\subsection{Procedure}
We split both event logs in a 2/3 training and 1/3 test set with a random process-instance-based sampling. As a baseline, we use the most cited next event PBPM technique from Tax et al.~\cite{tax.2017}. 
We evaluate the technique in two ways. 
First, we evaluate the optimisation of the KPI \textit{throughput time} (\textit{in-time} value) by the percentage of process instances that could comply with the temporal threshold for different prefix sizes. The temporal threshold is the average throughput time of a process instance in an event log.
Second, we evaluate the \textit{distance} from the ground truth process instance through the average Damerau-Levenshtein distance.
This metric determines the distance between two strings or sequences through the minimum number of operations (consisting of insertions, deletions or substitutions of a single character, or transposition of two adjacent characters), i.e. the lower the value, the more similar the strings are.

To train the multi-task LSTM $m_{pp}$, we apply the \emph{Nadam} optimisation algorithm %
with a \emph{categorical cross-entropy loss} for next activity predictions and a \emph{mean squared error} for \textit{throughput time} (KPI) predictions. 
Moreover, we set the batch size to 256, 
\ie gradients update after every 256\textsuperscript{th} sample of the training set. We set the default values for the other optimisation parameters.
For training the candidate selection model $m_{cs}$, we apply the nearest-neighbour-based ML algorithm ball tree~\cite{omohundro.1989}. Ball tree utilises a binary tree data structure for maintaining spatial data hierarchically. We choose a spatial-based algorithm to consider the semantic similarity between the suffixes of activities and KPI values.
Moreover, we set the hyperparameter $k$ (number of ``nearest" neighbours) of $m_{cs}$ to $5$, $10$ and $15$. Thereby, we check different sizes of the suffix candidate set.

Finally, technical details and the source code are available on GitHub\footnote{https://github.com/fau-is/next-best-action.}.

\subsection{Results}
\begin{figure}[htbp]
\centering
\includegraphics[width=\textwidth]{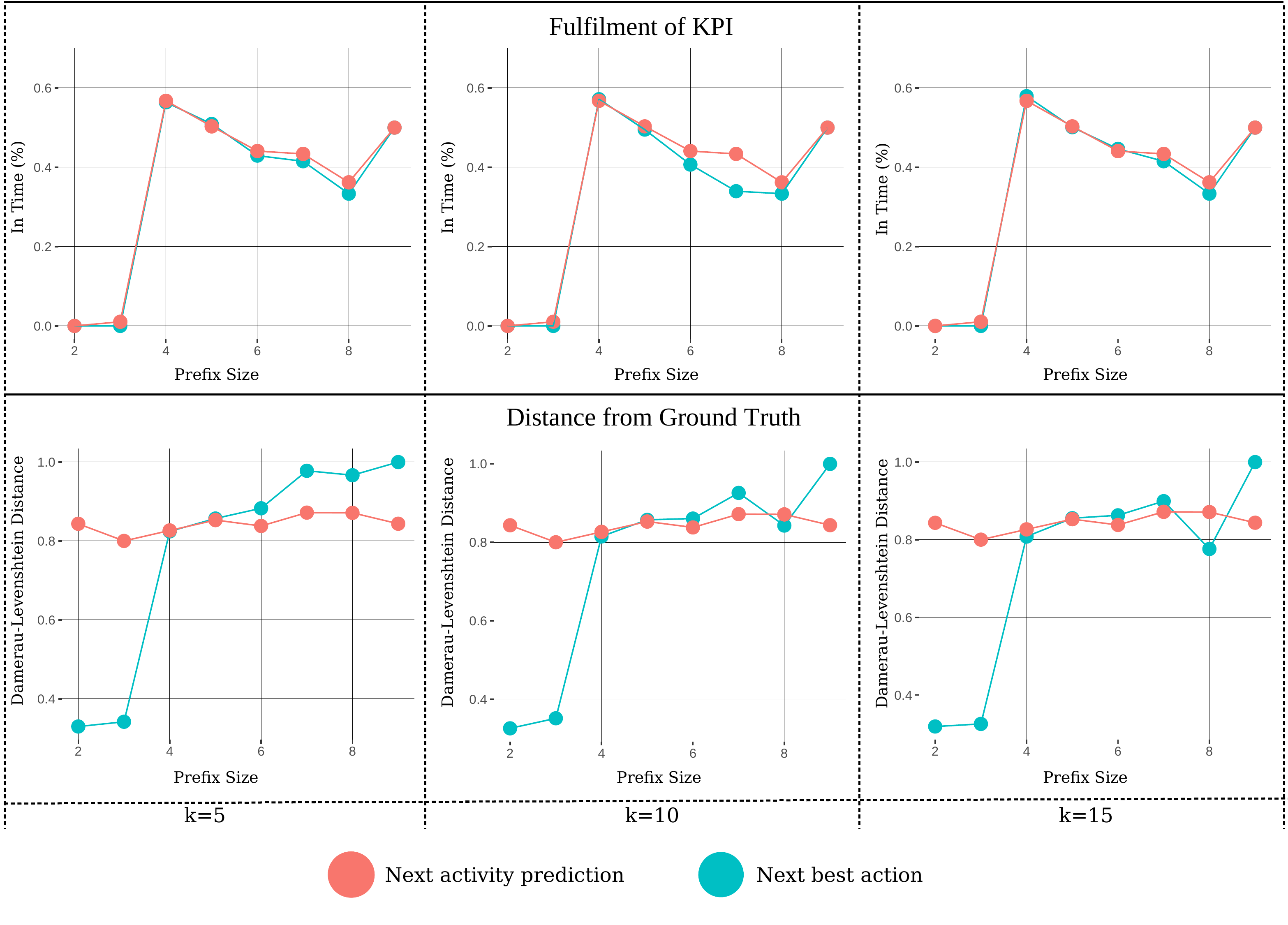}
\caption{Results for the \textit{helpdesk} event log.} 
\label{fig:res_hd}
\end{figure}

Fig.~\ref{fig:res_hd} shows the results for the \textit{helpdesk} event log.
\vspace{-0.3cm}
\vspace{-0.5cm}
For most of the prefixes in the \textit{helpdesk} event log, our technique's next best actions are more \textit{in time} than next activity predictions. 
While for $k=10$ next best actions have the lowest \textit{in-time} values compared to next activity predictions, \textit{in-time} values of next best actions with $k=5$ and $k=15$ are rather similar to each other.    
Furthermore, the higher the $k$, the lower is the \textit{distance} of the next best actions from the actual process instances.
Up to prefix size $4$, the \textit{distance} of the next best actions is lower compared to next activity predictions.

\begin{figure}[htbp!]
\centering
\includegraphics[width=\textwidth]{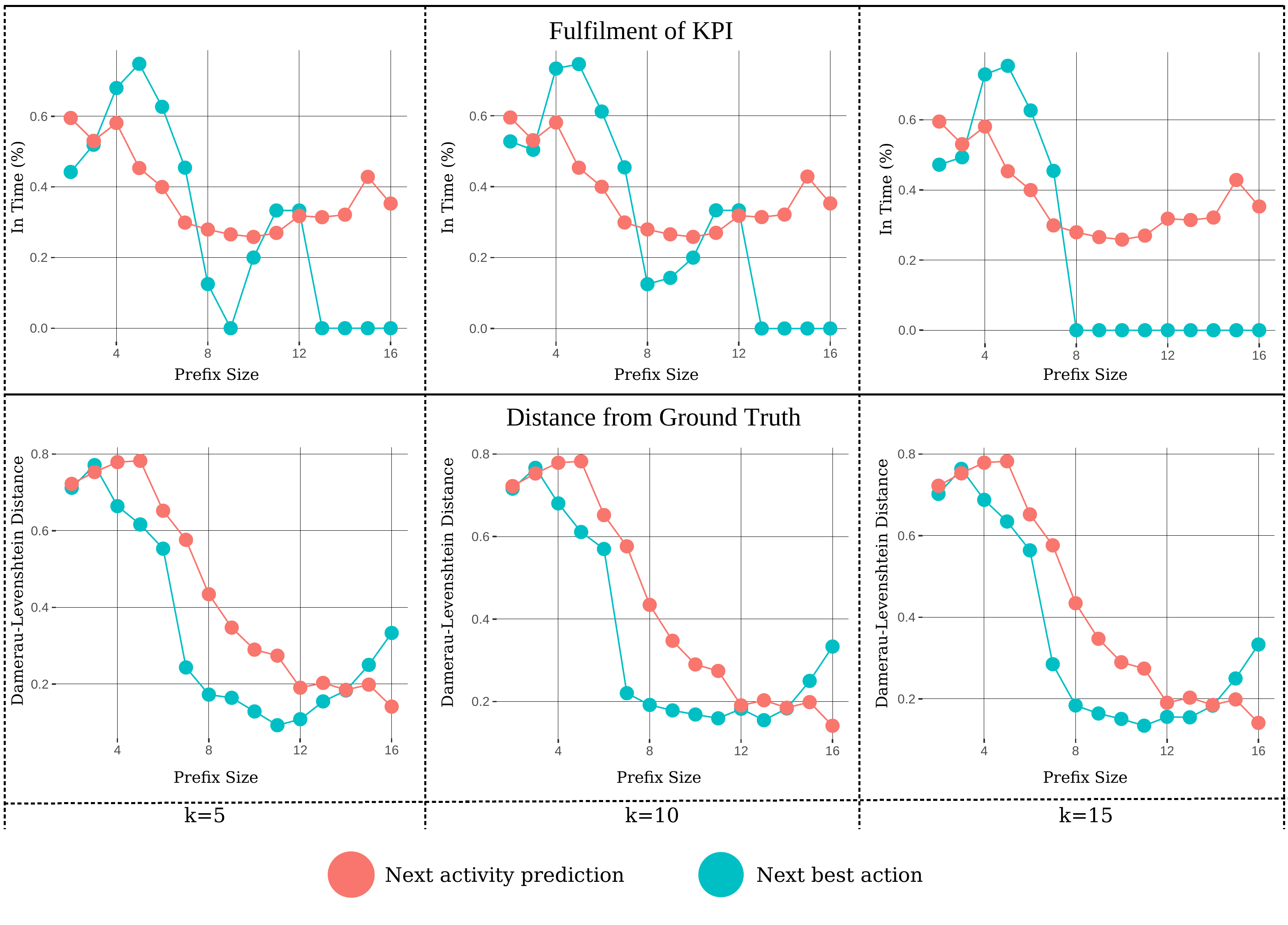}
\caption{Results for the bpi2019 event log.} 
\label{fig:res_bpi}
\end{figure}
Fig.~\ref{fig:res_bpi} shows the results for the \textit{bpi2019} event log.
For most of the prefixes with a size $\geq$ $8$, the next best actions of our technique are more \textit{in time} than next activity predictions. 
For $k=15$ and prefixes $\geq$ $8$, next best actions have an \textit{in-time} value of $0$.
With an increasing $k$, the \textit{in-time} values of the next best actions vary less from prefix size $2$ to $12$.    
In contrast to next activity predictions, the \textit{distance} of next best actions is lower for prefixes with a size $>3$ and $<15$.
Over the three $k$ values, the \textit{distance} of next best actions is rather similar.

\section{Discussion}
\label{sec:discussion}
Our contribution to academia and practice is a \magic{} technique that recommends the next best actions depending on a KPI while concerning BPS. 
Moreover, the evaluation presents an instantiation of our \magic{} technique. The KPI is the \textit{throughput time}, and a DCR graph realises BPS via the event log.

Based on our results, our \magic{} technique can provide actions with lower \textit{in-time} values and less \textit{distance} from the ground truth compared to the next most likely activities for both event logs.
However, the \textit{in-time} values (i.e. the percentage of process instances that could comply with the temporal threshold) of next best actions differs more from the baseline's next activity prediction for the \textit{bpi2019} event log than for the \textit{helpdesk} event log.
The \textit{helpdesk} event log has a lower instance variability than the \textit{bpi2019} event log. Therefore, fewer process paths exist from which our technique can recommend actions with lower \textit{in-time} values.
Further, the number of candidates $k$ has an effect on the KPI's optimisation.
While we get the actions with the lowest \textit{in-time} values with $k=10$ for the \textit{helpdesk} event log, the KPI values with $k=5$ and $k=15$ are similar to each other. For the \textit{bpi2019} event log, our technique provides actions with the lowest \textit{in-time} value if $k$ is set to $15$. The results with $k=5$ are similar to those of $k=10$.
A higher $k$ value leads to lower \textit{in-time} values in the \textit{bpi2019} event log because of a higher instance variability.
On the contrary, the \textit{helpdesk} event log needs a lower $k$ value.
Regarding the \textit{distance} from ground truth process instances, most of the next best actions (especially those for the \textit{bpi2019} event log) reach a better result than next activity predictions. 
A reason for that could be the limited predictive quality of the underlying DNN model for predicting next activities. However, our technique integrates control-flow knowledge and therefore overcomes this deficit to a certain degree.
Moreover, for the \textit{bpi2019} event log, our technique provides actions with lower \textit{in-time} values for prefixes with a size $\geq$ $8$.
In terms of the \textit{helpdesk} event log, we get actions with lower \textit{in-time} values for shorter prefixes. 
We suppose that our technique requires a longer prefix for event logs with higher instance variability to recommend next best actions. 
Finally, even though our technique provides actions with lower \textit{in-time} values, it seems that it does not terminate before the baseline. 
Our results show the aggregated values over different prefix sizes. 
Thus, we assume that few sequences, for which the termination can not be determined, distort the results. %

Despite all our efforts, our technique bears three shortcomings. 
First, we did not optimise the hyperparameters of the DNN model $m_{pp}$, e.g. via random search. Instead, we set the hyperparameters for $m_{pp}$ according to the work of Tax et al.~\cite{tax.2017}. We used the same setting since we compare our technique's next best actions to their next activity predictions.
Second, even though our technique is process-modelling-notation agnostic, we argue that declarative modelling is an appropriate approach for the process simulation. Due to its freedoms, declarative modelling facilitates the partial definition of the control-flow. 
As a consequence, we have a more flexible definition of a process's control-flow than by using a restricted procedural process model.
While our DNN-based technique copes well with rather flexible processes, other techniques using \textit{traditional} ML algorithms (e.g. a decision tree) might handle restricted processes faster and with a higher predictive quality.
Third, for our \magic{} technique's design, we neither consider cross-instance nor cross-business-process dependencies. In an organisational environment, additional effects like direct and indirect rebound effects can hinder our technique.

\section{Related Work}
\label{sec:relatedWork}
A variety of PBPM techniques were proposed by researchers as summarised by, e.g. Márquez-Chamorro et al.~\cite{marquez.2017} or Di Francescomarino et al.~\cite{di.2018}. 
Many of these techniques are geared to address the next activity prediction task. For that, most of the recent techniques rely on LSTMs~\cite{weinzierl.2020b} such as Weinzierl et al.~\cite{weinzierl.2020}.
To predict not only the next activities with a single predictive model, Tax et al.~\cite{tax.2017} suggest a multi-task LSTM-based DNN architecture. With this architecture, they predict the next activities and their timestamps. Metzger et al.~\cite{metzger.2019} extend their architecture by another LSTM layer to additionally predict the binary process outcome whether a delay occurs in the process or not. 
These techniques output predictions and do not %
recommend next best actions.

Furthermore, researchers suggested \magic{} approaches that raise alarms or recommend actions to prevent undesired activities. 
Metzger et al.~\cite{metzger2017predictive} investigate the effect of reliability estimates on (1) intervention costs (called adaption cost) and (2) the rate of non-violation of process instances by performing a simulation of a parameterised cost model. Thereby, they determine the reliability estimates based on the predictions of an ensemble of multi-layer perceptron classifiers at a pre-defined point in the process. In a later work~\cite{metzger2019dl}, reliability estimates were determined based on an ensemble of LSTM classifiers at different points in the process. The recommendation of actions is not part of these works.
Teinemaa et al.~\cite{teinemaa.2018} propose a concept of an alarm-based \magic{} framework. They suggest a cost function for generating alarms that trigger interventions to prevent an undesired outcome or mitigate its effect. In a later work~\cite{fahrenkrog.2019}, a multi-perspective extension of this framework was presented. In both versions, the framework focuses on alarms.
Gr{\"o}ger et al.~\cite{groger.2014} present a \magic{} technique that provides action recommendations for the next process step during the execution of a business process to avoid a predicted performance deviation. Performance deviation is interpreted as a binary outcome prediction, i.e. exists a deviation or not. In detail, an action recommendation comprises several action items and is represented by a rule extracted from a learned decision tree. An action item consists of the name and value of a process attribute. Even though this approach recommends actions in the form of process attribute values of the next process step which are optimised according to a KPI (e.g. lead time), process steps as next best actions are not recommended.  
Conforti et al.~\cite{conforti.2013} propose a \magic{} technique that predicts risks depending on the deviation of metrics during process execution. The technique's purpose is to provide decision support for certain actions such as the next process activity which minimises process risks. However, this technique can only recommend actions which are optimised regarding the KPI risk.   
Thus, with the best of our knowledge, there is no \magic{} approach that transforms next most likely activity predictions into the next best actions (represented by activities) depending on a given KPI.

\section{Conclusion}
\label{sec:conclusion}
Next activity predictions provided by PBPM techniques can be less beneficial for process stakeholders.
Based on our motivation and the identified research gap, we argue that there is a crucial need for a \magic{} technique that recommends the next best actions in running processes. 
We reached our research goal with the evaluation of our developed \magic{} technique in Sec. 5. 
Thereby, we show that our technique can outperform the baseline regarding KPI fulfilment and distance from ground truth process instances.
Further research might concern different directions. 
First, we plan to adapt existing loss functions for LSTMs predicting next most likely activities. Such a loss function can enable an LSTM to directly consider information on KPIs in the learning procedure.
Second, future research should further develop existing \magic{} approaches. More advanced multi-tasking DNN architectures can facilitate the recommendation of more sophisticated next best actions. For instance, next best actions that optimise more than one KPI.
Finally, we call for \magic{} techniques that are aware of concept evolution. Our technique is not able to recommend an activity as the best action if it was not observed in the training phase of the ML models. %

\section*{Acknowledgments}
This project is funded by the German Federal Ministry of Education and Research (BMBF) within the framework programme \textit{Software Campus} under the number 01IS17045.
 \bibliographystyle{splncs04}
 \bibliography{bpm}
\end{document}